# Design and Implementation of Computational Platform for Social-Humanoid Robot Lumen as an Exhibition Guide in Electrical Engineering Days 2015

Ahmad Syarif[#1], Ari Setijadi Prihatmanto[#2]

[#]School of Electrical Engineering and Informatics, Bandung Institute of Technology
Ganesha Street 10, Bandung 40132, Indonesia
[1]ahmadsyarif@students.itb.ac.id
[2]asetijadi@lskk.ee.itb.ac.id

*Abstract*— Social Robot Lumen is an Artificial Intelligence development project that aims to create an Artificial Intelligence (AI) which allows a humanoid robot to communicate with human being naturally. In this study, Lumen will be developed to be a tour guide in Electrical Engineering Days 2015 exhibition. In developing an AI, there are a lot of modules that need to be developed separately. To make the development easier, we need a computational platform which becomes basis for all developers to give easiness in developing the modules in parallel way. That computational platform that developed by the writer is called Lumen Server. Lumen Server has two main function, which are to be a bridge between all Lumen intelligence modules with NAO robot, and to be the communication bridge between those Lumen intelligence modules. For the second function, Lumen Server implements the AMQP protocol using RabbitMQ. Besides that, writer also developed a control system for robot movement called Lumen Motion. Lumen motion is implemented by modelling the movement of NAO robot and also by creating a control system using fuzzy logic controller. Writer also developed a program that connects all Lumen intelligence modules so that Lumen can act like a tour guide. The implementation of this program uses FSM and event-driven program. From implementation result, all the features which were designed are successfully implemented. By the developing of this computational platform, it can ease the development of Lumen in the future. For next development, it must be focused on creating integration system so that Lumen can be more responsive to the environment

## I. Introduction

The development of robotics recently has growing as popular field among researchers around the world. the very first concept of robotic is to create an automatic machine that can help human in industry. But today the concept of robotics has diversed even into social robot[2].

On this paper, the writer develops a computational platform to simplify the development of artificial intelligence (AI). This platform will become basis for other developers in developing Lumen intelligence modules to create AI that allows Lumen to act as tour guide in Electrical Engineering Days 2015 Exhibition.

Beside that, as additional, writer also developes a motion control system to control the movement of the robot NAO to get a better social ability. Writer also developes an integration system that integrates all Lumen intelligence modules that have been developed by other researchers in Lumen Project team, so that Lumen can act like a tour guide.

## II. Literature Study

### A. NAO Humanoid Robot

NAO is a brand for humanoid robot developed by Aldebaran Robotics[1]. It has 25 DOF, many sensors like camera, microphones, and embed with x86 AMD GEODE 500MHZ as its main computer. NAO has modified Linux Gentoo OS called openNAO which run an application called NAOqi. NAOqi is a programming framework that allow the developer to developer NAO easier.

NAOqi has an application programming interface (API) that allow developer to get data from NAO sensors as well as control the actuators. The API is available in many programming languages such as C++, Python, .NET, Java, and MATLAB.

### B. AMQP and RabbitMQ

AMQP stands for Advance Message Queueing Protocol. It's a standard protocol in application layer used by message-oriented middleware[5]. It allows messaging between many client or computer with simple procedure and many features.

RabbitMQ is on of message-oriented middleware that implements AMQP[6]. It's an open source program that can run in almost all famous operating system. It also has great API to create a client application. The API is available in almost famous programming languages, such as C++, Python, and .NET

### C. Fuzzy Logic Controller

Fuzzy Logic Controller (FLC) is a controller that uses fuzzy set and fuzzy logic[3][4]. Fuzzy set is a set of number that has no crips value, whereas fuzzy logic is mathematical

logic that uses fuzzy set. Every fuzzy set can be decribed with this fuction

$$A = \{(x, \mu_A(x)) | x \in X\} \quad (1)$$

With $\mu_A(x)$ is membership function (MF) that can be defined using many equation. On of them is gaussian MF which written as

$$\mu_A(x) = gaussian(x; c, \sigma) = e^{-\frac{1}{2}\left(\frac{x-c}{\sigma}\right)^2} \quad (2)$$

The fuzzy logic implement rule, one of which is if-the rules that can be written with a simple expression using linguistic value and variable such as "if the temperature is hot, set the AC to be cold". Temperature is a linguistic variable and hot and cold are linguistic value

### III. DESIGN AND IMPLEMENTATION

In this chapter, we will show the designs and implementation of each modules by first explaining about the complete architecture of Lumen.

#### A. Lumen Architecture

Lumen is a project done by number of researcher. Thus we need an architecture that become basis in develoment

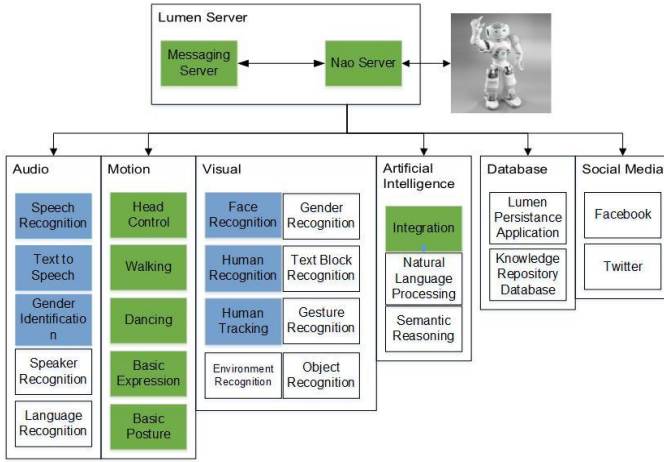

Fig. 1 Lumen Architecture

There are basically 7 main modules which are server, audio, visual, motion, AI, database, and social media. In this paper, we design the software for server, motion, and AI modules. we separate the explanation of the design for each module

#### B. Lumen Server

Lumen Server is the main topic in this paper. It has two main functions which as a bridge from the software to NAO robot and as communication bridge for all other modules. from these function, we design separate software in order to achieve the functionality

The first function is manifested with submodule called NAO Server. The main task of this submodule is to get real-time data from NAO sensor and to control the NAO motor actuator. The design of the software shown in Fig. 2 and Fig. 3 below

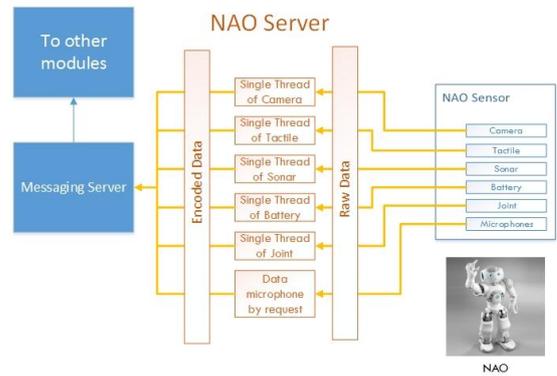

Fig. 2 Data Aquisition from NAO sensor

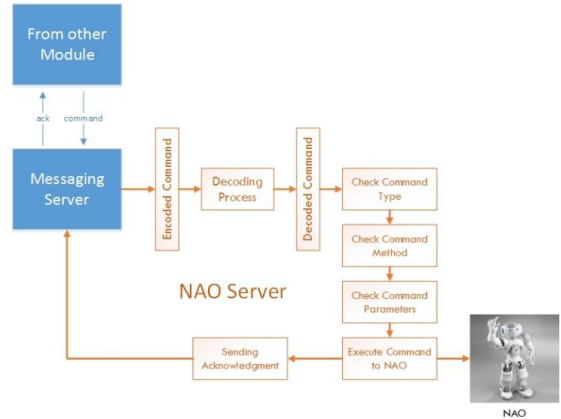

Fig. 3 Command execution to NAO actuator

To achieve real-time data aquisition and command handling, we implemented multithreading algorithm and set a single thread for each type of data and for command handling.

For the second function, we used RabbitMQ server as the messaging server and then design the routing key for each module and the JSON data format for each data and message. The list of routing keys are show in Fig.4 to Fig. 7 below

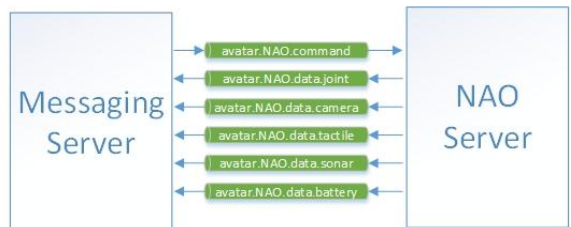

Fig. 4 Routing key for NAO Server

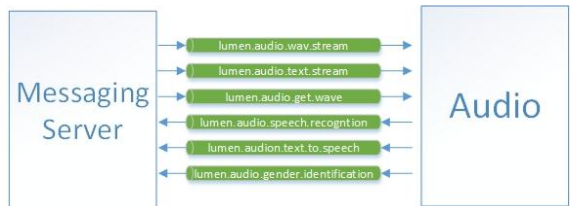

Fig. 5 Routing key for Audio Module

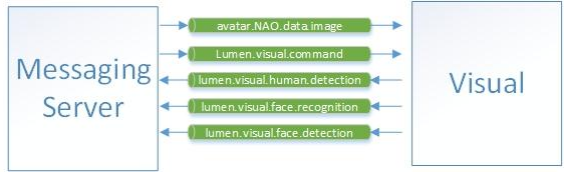

Fig. 6 Routing key for Visual Module

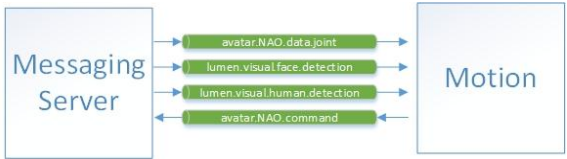

Fig. 7 Routing key for Motion Module

If any new module which want to connect to the platform, its routing key must be started either with 'avatar' or 'lumen'. 'avatar' is used for modules that deal with NAO or other robot or virtual robot platform, while 'lumen' is used for modules that deal with Lumen intelligence modules.

### C. Lumen Motion

Lumen Motion is responsible to handle NAO movement to create natural body language for communication. Writer designed three different motion controller for Lumen Motion.

The first is Head Control. This submodul controls NAO head to make sure NAO always face toward the visitor. In order to achive that, we used FLC in designing the controller

We use Mamdani Fuzzy Model, we define four linguistic variables, those are FaceXLoc, FaceYLoc, AngleX, and AngleY. Each linguistic variable has three linguistic values those are Negative, Zero, and Positive. Each linguistic value is defined using gaussian MF.

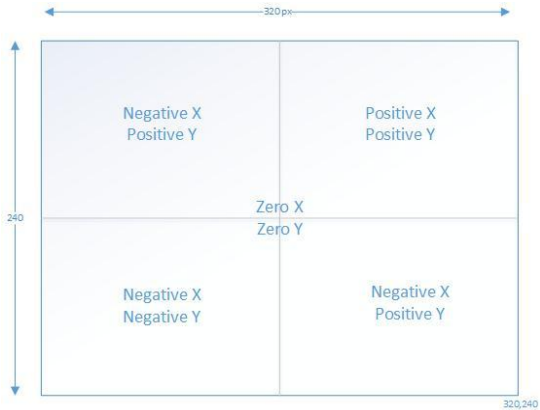

Fig. 8 Linguistic value of face location

After that, we define six rules which control the output of the controller as shown in Table 1 below

Tabel 1 List of FLC rules

| Antecedent | Consequnce |
|---|---|
| FaceXLoc is negative | AngleX is positive |
| FaceXLoc is positive | AngleX is negative |
| FaceXLoc is zero | AngleX is zero |
| FaceYLoc is negative | AngleY is positive |
| FaceYLoc is positive | AngleY is negative |
| FaceYLoc is zero | AngleY is zero |

The fuzzy operation that we use is min-max operation. For deffuzification, we use centroid method in order to get crips value from fuzzy set which become the input for NAO headyaw and headpitch joint. The membership function of the linguistic variables are defined below

$$negative = gaussian(x; c_n, \sigma_n) = e^{-\frac{1}{2}\left(\frac{x-c_n}{\sigma_n}\right)^2}, x \in X \quad (3)$$

$$positive = gaussian(x; c_p, \sigma_p) = e^{-\frac{1}{2}\left(\frac{x-c_p}{\sigma_p}\right)^2}, x \in X \quad (4)$$

$$zero = gaussian(x; c_z, \sigma_z) = e^{-\frac{1}{2}\left(\frac{x-c_z}{\sigma_z}\right)^2}, x \in X \quad (5)$$

The value of each parameter in equation (3) to equation (5) are defined in table below

Table 2 Value of paremeters

| Parameter | FaceXLoc | FaceYLoc | AngleX | AngleY |
|---|---|---|---|---|
| $c_n$ | 0 | 0 | -15 | -7 |
| $\sigma_n$ | 80 | 70 | 10 | 6 |
| $c_p$ | 320 | 120 | 0 | 0 |
| $\sigma_p$ | 80 | 40 | 10 | 6 |
| $c_z$ | 160 | 240 | 15 | 7 |
| $\sigma_z$ | 50 | 70 | 10 | 6 |

For the walking and basic postures, we implemented it using API from NAOqi. The codes we used are shown below

```
MotionProxy motion = new MotionProxy(NAO_IP,
NAO_PORT);
motion.moveInit();
motion.moveTo(0.2f,0.0,0.0);

RobotPostureProxy posture = new
RobotPostureProxy(NAO_IP, NAO_PORT);
posture.goToPosture("Stand",0.5f);
```

we put the codes in NAO Server so that other modules can access it through sending the right message to NAO Server.

For dancing, singing movement, and hand waving, we first design the angle of each joint to create the particular movement. That we implement the codes in NAO Server as 'dancing', 'singing', and 'goodbye' methods that can be access by other module by sending particular message to NAO Server.

For dancing, we design NAO to dance 'gangnam style' dance. For singing, we design NAO to sing 'manuk jajali' song while waving its hand and pretending to hold a microphone

### D. Lumen AI Integration

Lumen AI Integration is a module that connect all Lumen intelligence modules to create the behavior for Lumen to be a tour guide. The architecture for the program is shown in Fig. 8 below

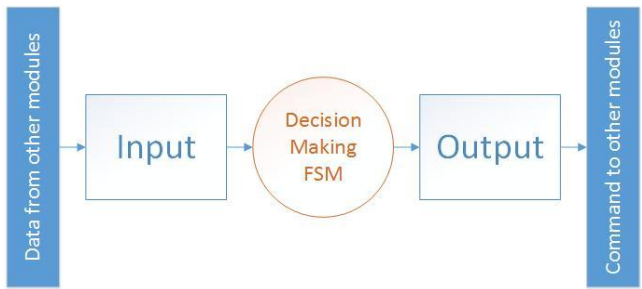

Fig. 8 AI Integration Architecture

We design a FSM to control the state of Lumen so that Lumen can take a decision based on the condition. The FSM has 15 state with condition as shown in Fig. 9 which is originally design from Audio Module

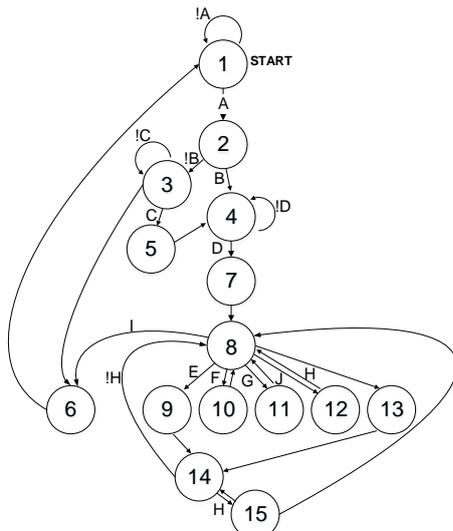

Fig.9 AI FSM

## IV. RESULT AND ANALYSIS

The result of the implementation will be explaining separately in each part of this section.

### A. Lumen Server

From the implementation, NAO Server is able to get data from five NAO Sensor in paralel. Not only that, while getting data from NAO, NAO Server is also able to execute command to NAO.

Fig. 10 time elapsed to get image from NAO

Fig. 11 NAO Server executes command to NAO

For the messaging server, since we implement is using RabbitMQ, no flaw that has been shown during the implementation of the system. That fact that other modules are able to get data from NAO Server are the prove that messaging server work perfectly

### B. Lumen Motion

This Fig. 12 below shows the head control program of NAO

Fig. 12 angle and face location

As show in the Fig. 12, the program set the angle of NAO head joint based on the location of the face in the frame. Fig. 13 shows the face initial position

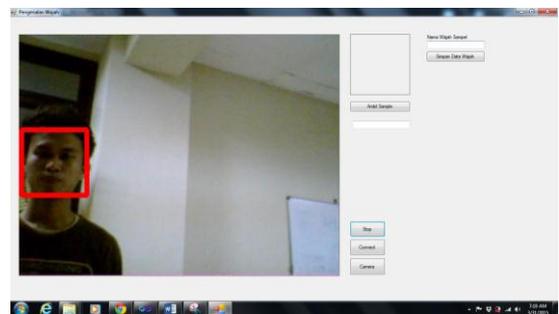

Fig. 13 face initial position

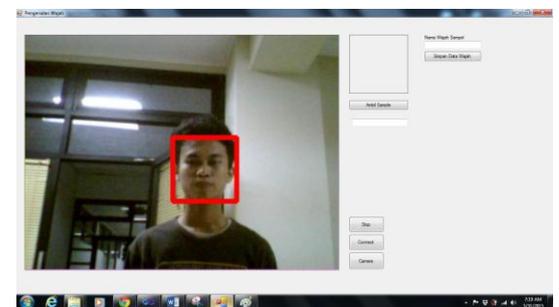

Fig. 14 face final position

And Fig.14 shows the final position, from this result we see that controller is able to control NAO head toward face

For walking and basic posture, here are the pictures that show the result

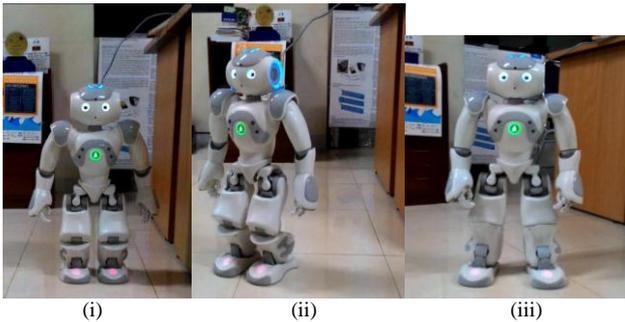

(i) (ii) (iii)
Fig. 15 NAO perform walking from (i) initial position to (ii) final position, and (iii) stand posture

The result for dancing, singing, and waving hand are shown in figures below

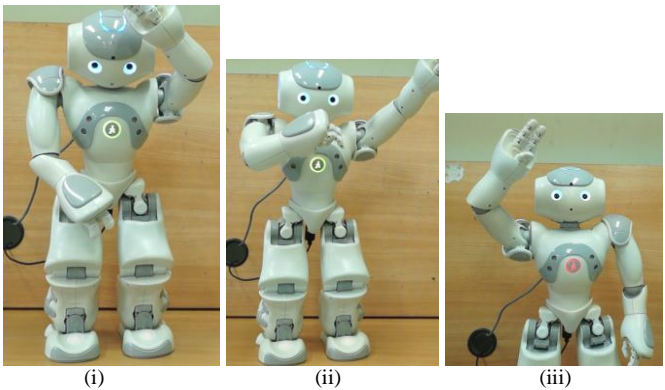

(i) (ii) (iii)
Fig. 16 (i) dancing, (ii) singing, and (iii) hand waving

*C. Lumen AI Integration*

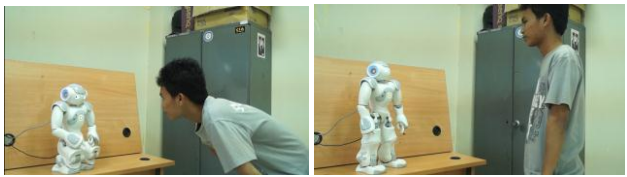

Fig. 17 NAO is interacting with human

V. CONCLUTION

The figures show the result of Lumen AI Integreation implementation. The robot is able to perform the state machine and thus interact with human as tour guide

1. Lumen Server is able to become bridge between for all modules and NAO, as well for all modules to other modules.
2. FLC is able to control the heading of NAO head toward the face in the frame
3. NAO is able to perform walking, six basic postures, dancing, singing movement, and hand waving
4. The designed FSM is able to control NAO state while interacting with human. The more complex the FSM be, Lumen can act more naturally.

ACKNOWLEDGMENT

Thanks to the all member of Lumen Project for helping this research, especially for my supervisor for giving me many advice so that this research is going well. This research is fully supported by Control System and Computer Laboratory, Bandung Institute of Technology

# Perancangan dan Implementasi Platform Komputasi untuk Robot Sosial Humanoid Lumen Sebagai Pemandu Pameran pada ElectricalEngineering Days 2015

Ahmad Syarif[#1], Ari Setijadi Prihatmanto[#2]

[#]*Sekolah Teknik Elektro dan Informatika, Institut Teknologi Bandung*
*Jalan Ganesha 10, Bandung 40132, Indonesia*
[1]ahmadsyarif@students.itb.ac.id
[2]asetijadi@lskk.ee.itb.ac.id

*Abstrak*— Sosial Robot Lumen adalah proyek pengembangan kecerdasan buatan yang bertujuan untuk menciptakan kecerdasan buatan atau *artificial intelligence* (AI) yang memungkinkan robot untuk dapat berkomunikasi dengan manusia secara alami. Pada studi ini, Lumen dikembangkan untuk menjadi pemandu pada pameran Electrical Engineering Days 2015. Dalam pengembangan AI, ada banyak sekali modul-modul yang harus dikembangkan secara terpisah. Untuk memudahkan pengembangan, perlu ada suatu platform komputasi yang menjadi landasan bagi setiap pengembang serta memudahkan pengembangan secara paralel terhadap modul modul ini. Platform komputasi ini kemudian disebut Lumen Server. Lumen Server berfungsi sebagai jembatan antara modul kecerdasan Lumen dan robot NAO sebagai platform robot yang digunakan dalam pengembangan Lumen. Selain itu, Lumen Server juga berfungsi untuk menjadi jembatan komunikasi antara modul-modul kecerdasan Lumen. Untuk fungsi kedua ini, Lumen Server mengimplementasikan AMQP dengan menggunakan RabbitMQ. Selain itu, sebagai tambahan, penulis juga mengembangkan suatu kendali gerakan yang disebut Lumen Motion. Lumen Motion diimplementasikan dengan memodelkan gerakan NAO serta membuat sistem kendali dengan Fuzzy Logic Controller. Penulis juga mengembangkan suatu program yang menghubungkan semua modul kecerdasan Lumen sehingga Lumen dapat bertindak layaknya pemandu pada pameran. Implementasi program menggunakan FSM serta event-driven program. Dari hasil implementasi, semua fitur yang dirancang dapat diimplementasikan dengan baik. Diharapkan dengan pembuatan platform komputasi ini, dapat memudahkan pengembangan Lumen kedepannya. Untuk pengembangan selanjutnya, lebih ditekankan pada pembuatan sistem integrasi yang lebih terfokus sehingga Lumen dapat lebih responsif terhadap kondisi sekitar.

Kata kunci — robot, humanoid, server, RabbitMQ, Fuzzy Logic Controller, AI.

I. INTRODUCTION

Dunia robotika telah berkembang menjadi salah satu bidang yang paling diminnati diantara para peneliti di dunia. Konsep awal robot, yaitu mesin otomatis yang dapat membantu manusia di bidang industry. Namun, sekarang konsep robot telah berkembang hingga robot sosial[2].

Pada karya tulis ini, penulis mengembangkan suatu platform komputasi untuk pengembangan *artificial intelligence* (AI). Platform ini akan menjadi dasar bagi setiap pengembang dalam mengembangkan modul kecerdasan Lumen untuk membuat suatu AI yang dapat membuat Lumen menjadi pemandu pameran pada *Electrical Engineering Days 2015*

Selain itu, sebagai tambahan, penulis juga mengembangkan sebuah sistem kendali gerak untuk mengendalikan gerakan NAO untuk menciptakan kemampuan sosial yang lebih baik. Penulis juga mengembangkan suatu sistem integrasi yang mengintegrasikan semua modul kecerdasan Lumen yang dikembangkan oleh pengembang lain, sehingga Lumen dapat menjadi pemandu pameran

II. STUDI PUSTAKA

A. *Robot Humanoid NAO*

NAO adalah suatu mrek untuk robot humanoid yang dikembangkan oleh Aldebaran Robotics[1]. NAO memiliki 25 DOF, beberapa sensor seperti kamera, *microphones*. NAO juga memiliki x86 AMD GEODE 500MHz sebagai komputer utama. NAO memiliki OS modifiikasi dari Linux Gentoo bernama openNAO yang menjalankan suatu program bernama NAOqi. NAOqi adalah *framework* pemrograman yang memudahkan pengembangan NAO.

NAOqi memiliki sebuah *application programming interface* (API) yang memungkinkan pengembangan mengakusisi data dari sensor NAO maupun mengendalikan aktuator NAO. API tersebut tersedia dalam beberapa bahasa pemrogramanan seperti C++, Python, .NET, Java, dan MATLAB

B. *AMQP dan RabbitMQ*

AMQP adalah singkatan dari *advance message queueing protocol*. AMPQ adalah protokol standar pada *application layer* yang digunakan oleh *message-oriented middleware*[5]. AMQP memungkinkan proses pengiriman pesan antara banyak *client* atau komputer dengan prosedur sederhana dan banyak fitur.

RabbitMQ adalah salah satu *message-oriented middleware* yang menerapkan AMQP[6]. RabbitMQ adalah program *open source* yang dapat dijalankan di hampir semua sistem operasi. RabbitMQ juga memiliki API untuk pengembangan program *client* yang tersedia dalam beberapa bahasa seperti C++, Python, dan .NET

## C. Fuzzy Logic Controller

*Fuzzy Logic Controller* (FLC) adalah suatu pengendali yang menggunakan himpunan samar dan logika samar[1][2]. Himpunan samar adalah suatu himpunan bilangan yang tidak memiliki batas tegas, sementara logika samar adalah logika yang matematika menggunakan himpunan samar. Setiap himpunan samar dapat dituliskan dengan fungsi berikut

$$A = \{(x, \mu_A(x))|x \in X\} \quad (1)$$

Dengan $\mu_A(x)$ adalah suatu fungsi keanggotaan (MF) yang dapat dituliskan dalam banyak persamaan, diantaranya gaussian MF

$$\mu_A(x) = gaussian(x; c, \sigma) = e^{-\frac{1}{2}\left(\frac{x-c}{\sigma}\right)^2} \quad (2)$$

Logika samar memiliki aturan, salah satu nya adalah aturan jika-maka yang dapat dituliskan dengan variabel dan nilai linguistik. Contohnya, "jika suhu panas, maka AC diatur agar dingin". Suhu adalah variabel linguistik sementara panas dan dingin adalah nilai linguistik

### III. PERANCANGAN DAN IMPLEMENTASI

Pada bagian ini, akan ditunjukan rancangan dan implementasi dari tiap modul dengan terlebih dahulu menjelaskan.

### A. Arsitektur Lumen

Lumen adalah proyek yang dijalankan oleh banyak peneliti. Untuk itu dibutuhkan suatu arsitektur yang menjadi dasar pengembangan seperti terlihat pada Gambar 1

Terdapat 7 buah modul utama, yaitu server, audio, visual, motion, AI, database, dan media sosial. Pada karya ilmiah ini, penulis mengembangkan *software* untuk server, motion, dan AI. Rancangan untuk ketiganya akan dijelaskan secara terpisah

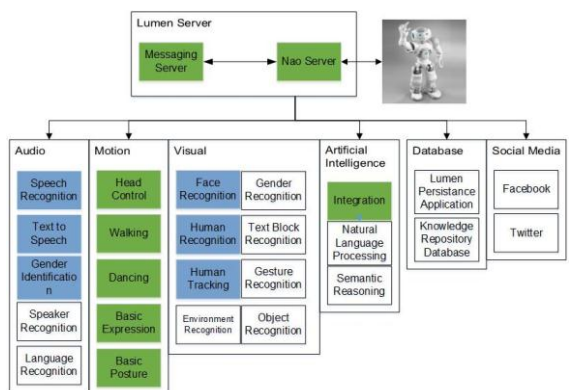

Gambar. 1 Arsitektur Lumen

### B. Lumen Server

Lumen Server adalah topik utama pada karya ilmiah ini. Lumen server memiliki dua fungsi utama yaitu sebagai jembatan komunikasi antara program dan NAO serta jembatan komunikasi antara tiap modul. Dari dua tujuan itu, dikembangkan dua jenis *software*

Tujuan pertama dimanifestasikan dengan submodul bernama NAO Server. Tugas utamanya adalah mengakusisi data secara *real-time* dari sensor NAO dan mengendalikan aktuator NAO. rancangan dari *software*-nya dapat dilihat pada Gambar 2 dan Gambar 3.

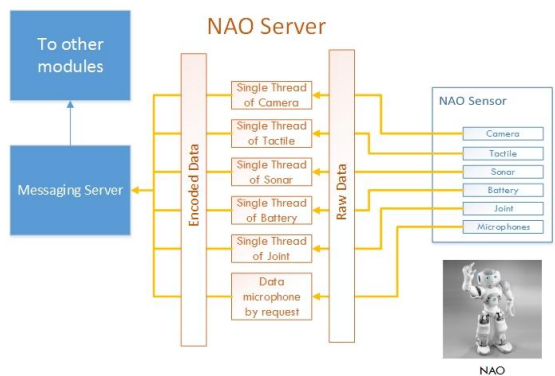

Gambar 2 Akuisisi data dari sensor NAO

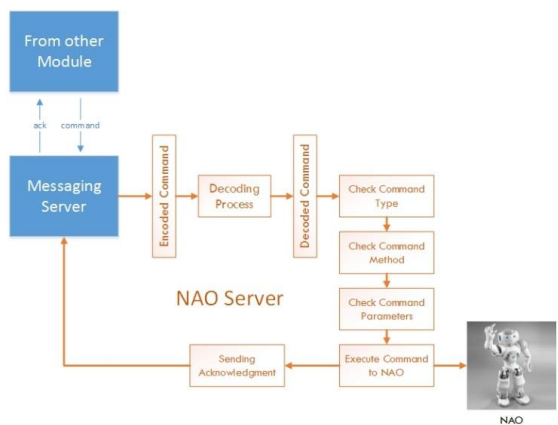

Gambar 3 Eksekusi perintah ke NAO

Untuk membuat sistem akuisisi data dan eksekusi perintah secara *real-time*, digunakan algoritma *multithreading* dan mengatur satu buah *thread* untuk akuisisi tiap jenis data dan eksekusi perintah.

Untuk fungsi kedua, digunakan RabbitMQ Server sebagai *Messaging Server,* lalu dirancang *routing key* untuk tiap modul dan juga format data JSON untuk tiap data dan pesan. Daftar *routing key* ditunjukkan oleh Gambar 4 hingga Gambar 7.

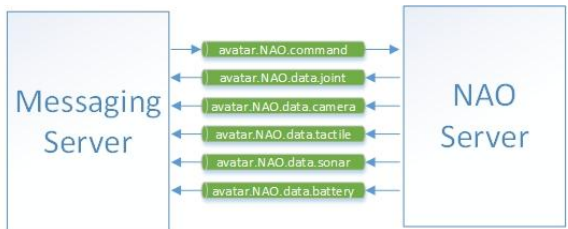

Gambar 4 Routing key untuk NAO Server

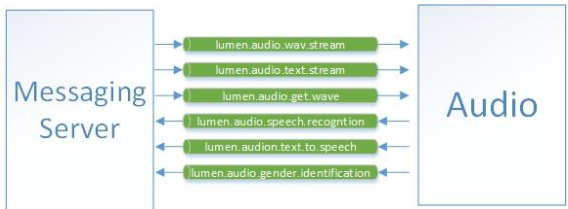

Gambar 5 Routing key untuk modul Audio

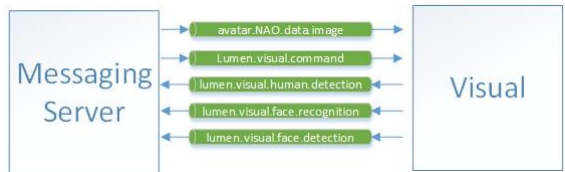

Gambar 6 Routing key untuk modul Visual

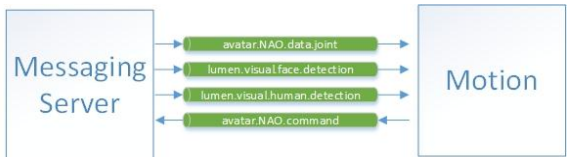

Gambar 7 Routing key untuk modul Motion

Jika ada modul baru yang ingin terkoneksi dengan platform ini, maka *routing key* –nya harus dimulai dengan 'avatar' atau 'lumen'. 'avatar' digunakan oleh modul yang berhubungan dengan NAO atau robot lainnya, sementara 'lumen' digunakan oleh modul kecerdasan Lumen.

*C. Lumen Motion*

Lumen Motion bertanggung jawab untuk mengendalikan gerakan NAO untuk menciptakan bahasa tubuh yang alami. Penulis merancang tiga jenis pengendali untuk Lumen Motion.

Yang pertama adalah *Head Control*. Submodul ini mengendalikan kepala NAO agar selalu menghadap ke pengunjung. Untuk itu, penulis menggunakan FLC dalam merancangan pengendali.

Penulis menggunakan *Mamdani Fuzzy Model* dan mendefenisikan empat buah variabel linguistik, yaitu FaceXLoc, FaceYLoc, AngleX, dan AngleY. Setiap variabel memiliki tiga nilai linguistik yaitu Negative, Zero, dan Positive yang didefenisikan menggunakan Gaussian MF

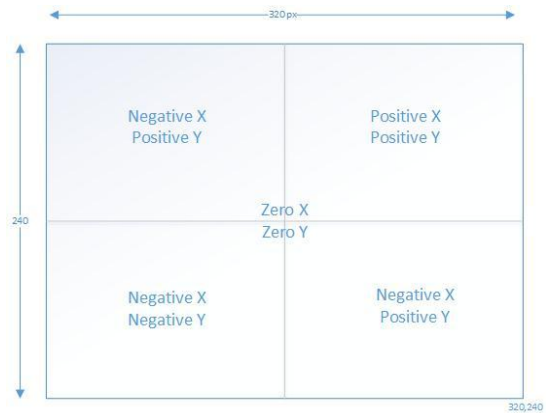

Gambar 8 Nilai linguistik untuk lokasi wajah

Setelah itu, didefenisikan enam buah aturan yang mengendalikan output dari pengendal seperti terlihat pada Tabel 1.

Tabel 1 List of FLC rules

| Antecedent | Consequnce |
|---|---|
| FaceXLoc negative | AngleX positive |
| FaceXLoc positive | AngleX negative |
| FaceXLoc zero | AngleX zero |
| FaceYLoc negative | AngleY positive |
| FaceYLoc positive | AngleY negative |
| FaceYLoc zero | AngleY zero |

Operasi samar yang digunakan adalah operasi min-maz. Untuk defuzifikasi, digunakan metode *centroid* untuk mengambil nilai tegas yang menjadi input untuk sendiri HeadYaw dan HeadPitch NAO. *Membership function* dari variabel linguistik diatas didefenisikan dengan persamaan berikut

$$\text{negative} = \text{gaussian}(x; c_n, \sigma_n) = e^{-\frac{1}{2}\left(\frac{x-c_n}{\sigma_n}\right)^2}, x \in X \quad (3)$$

$$\text{positive} = \text{gaussian}(x; c_p, \sigma_p) = e^{-\frac{1}{2}\left(\frac{x-c_p}{\sigma_p}\right)^2}, x \in X \quad (4)$$

$$\text{zero} = \text{gaussian}(x; c_z, \sigma_z) = e^{-\frac{1}{2}\left(\frac{x-c_z}{\sigma_z}\right)^2}, x \in X \quad (5)$$

Nilai untuk tiap parameter pada persamaan (3) hingga persamaan (5) didefenisikan pada tabel dibawah ini

Tabel 2 Nilai Parameter

| Parameter | FaceXLoc | FaceYLoc | AngleX | AngleY |
|---|---|---|---|---|
| $c_n$ | 0 | 0 | -15 | -7 |
| $\sigma_n$ | 80 | 70 | 10 | 6 |
| $c_p$ | 320 | 120 | 0 | 0 |
| $\sigma_p$ | 80 | 40 | 10 | 6 |
| $c_z$ | 160 | 240 | 15 | 7 |
| $\sigma_z$ | 50 | 70 | 10 | 6 |

Untuk *walking* dan *Basic Postures*, digunakan API dari NAOqi. Kode program yang digunakan dapat dilihat dibawah ini

```
MotionProxy motion = new MotionProxy(NAO_IP,
NAO_PORT);
motion.moveInit();
motion.moveTo(0.2f,0.0,0.0);

RobotPostureProxy posture = new
RobotPostureProxy(NAO_IP, NAO_PORT);
posture.goToPosture("Stand",0.5f);
```

Kode tersebut ditempatkan di NAO Server sehingga setiap modul dapat mengaksesnya dengan mengirim pesan yang sesuai ke NAO Server

Untuk *Dancing, Singing Movement,* dan *Hand Waving*, pertama dirancang sudut untuk tiap sendi untuk menciptakan gerakan tertentu. Setelah itu rancangan tersebut diimplementasikan pada NAO server sebagai *method* 'dancing', 'singing', dan 'goodbye' yang dapat diakses oleh modul lain dengan mengirimkan pesan sesuai ke NAO Server.

Untuk *Dancing*, penulis merancangan NAO untuk menarikan tarian 'gangnam style'. Untuk *Singing*, penulis merangan NAO menyanyikan lagu 'manuk jajali' sambil melambaikan tangan dan seakan-akan memegang *microphone*

### D. Lumen Integration

Modul Lumen AI Integration adalah modul yang menyambungkan seluruh modul kecerdasan Lumen untuk menciptakan prilaku Lumen sebagai pemandu pameran. Arsitektur dari programnya ditunjukkan oleh Gambar 8 dibawah ini

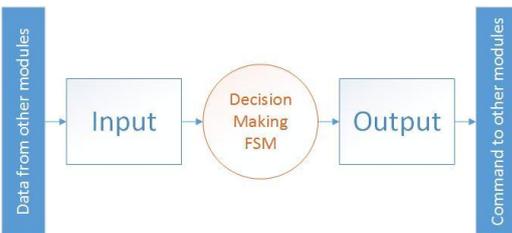

Gambar 8 Arsitektur AI Integration

Sebuah FSM dirancang sebagai pengendali *state* dari Lumen sehingga Lumen dapat mengambil keputusan sesuai kondisi. FSM ini memiliki 15 buah *state* dengan kondisi seperti ditunjukkan Gambar 9 dan Tabel 2 berikut

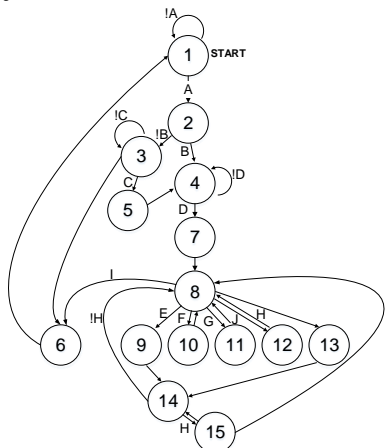

Gambar 9 AI FSM

## IV. HASIL DAN ANALISIS

Hasil dari implementasi akan dijelaskan pada bagian terpisah di bab ini.

### A. Lumen Server

Dari hasil implementasi, NAO Server mampu mengambil lima jenis data sensor NAO secara paralel dan mengeksekusi perintah secara bersamaan

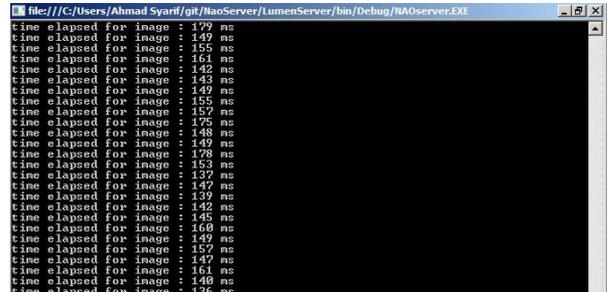

Gambar 10 waktu pengambilan data gambar dari kamera NAO

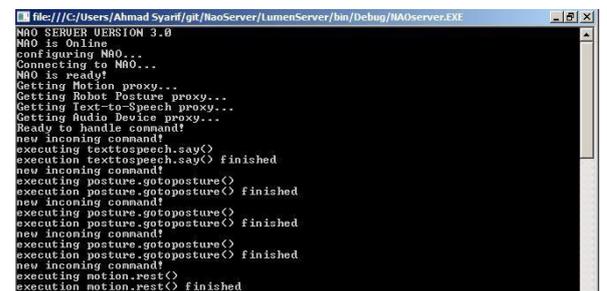

Gambar 11 NAO Server mengeksekusi perintah keNAO

Untuk *Messaging Server*, karena kita menggunakan RabbitMQ, tidak terdapat kelemahan selama implementasi. Fakta bahwa semua modul dapat mengambil data dari NAO Server adalah bukti bahwa *Messaging Server* bekerja dengan baik.

### B. Lumen Motion

Gambar 12 dibawah menunjukkan program *Head Control* untuk NAO

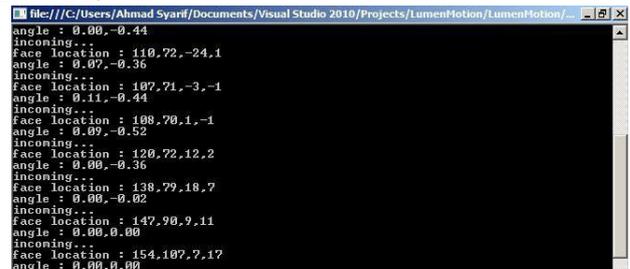

Gambar 12 Sudut and dan lokasi wajah

Seperti ditunjukkan oleh Gambar 12, program ini mengatur sendi kepala NAO berdasarkan posisi wajah di *frame*. Gambar 13 menunjukkan posisi awal dari wajah

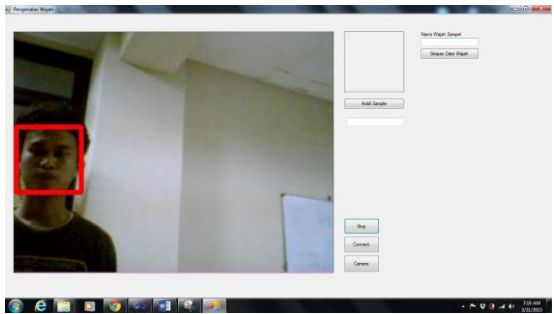
Gambar 13 Lokasi awal wajah

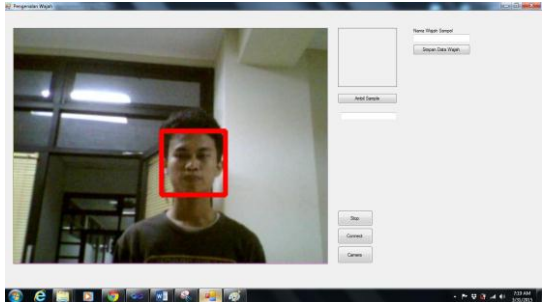
Gambar 14 Lokasi akhir wajah

Gambar 14 menunjukkan lokasi akhir wajah. Dari hasil ini, dapat dilihat bahwa pengendali mampu mengendalikan kepala NAO menghadap ke wajah

Untuk *walking* dan *basic postures*, berikut adalah gambar yang menunjukkan hasil implementasi

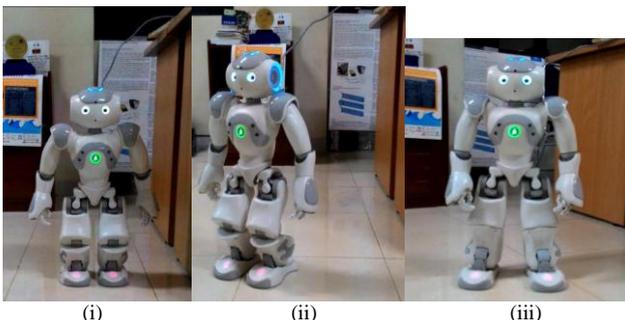
(i)    (ii)    (iii)
Gambar 15 NAO berjalan dari (i) posisi awal ke (ii) posisi akhir, dan (iii) melakukan posture *stand*

Hasil dari implementasi *dancing, singing,* dan *hand waving* di tunjukkan oleh gambar dibawah ini

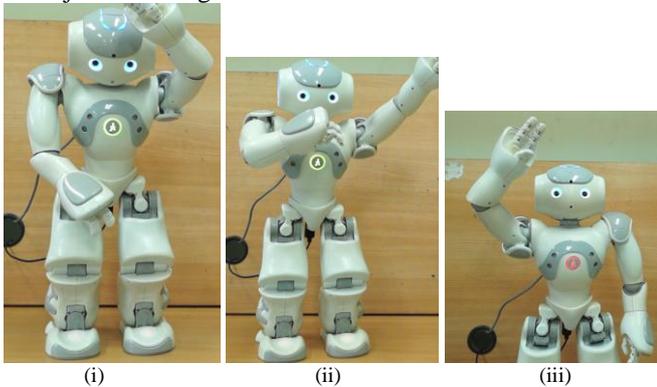
(i)    (ii)    (iii)
Gambar 16 (i) dancing, (ii) singing, and (iii) hand waving

*C. Lumen AI Integration*

Gambar dibawah ini menujukkan hasil implementasi Lumen AI Integration. Robot mampu memperagakan FSM sehingga dapat berinteraksi dengan manusia sebagai pemandu pameran

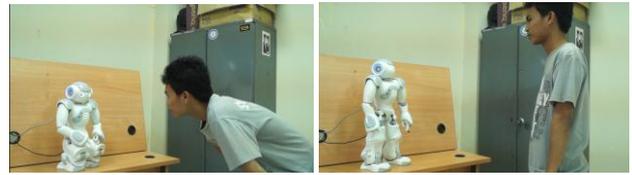
Gambar 17 NAO sedang berinteraksi dengan manusia

## V. CONCLUSIONS

1. Lumen Server mampu menjadi jembatan komunikasi antara semua modul dan NAO, serta antara modul satu dan lainnya
2. FLC mampu megendalikan kepala NAO menghadap ke wajah di *frame*
3. NAO mampu menjalankan *walking*, enam *basic postures, dancing, singing movement,* dan *hand waving*
4. Rancangan FSM mampu mengendalikan *state* Lumen sambil berinteraksi dengna manusia. Semakin kompleks FSM nya, maka Lumen akan semakin alami dalam beriteraksi

ACKNOWLEDGMENT

Terima kasih kepada semua anggota tim proyek Lumen yang telah membantu penelitian ini, khususnya kepada pembimbing yang telah memberikan banyak saran dan masukan untuk riset ini. Penelitian ini didukung penuh oleh Laoratorium Sistem Kendali dan Komputer LSKK ITB.